\pdfoutput=1
\PassOptionsToPackage{table}{xcolor}
\documentclass[11pt]{article}

\usepackage[final]{acl}

\usepackage{times}
\usepackage{latexsym}

\usepackage[T1]{fontenc}
\usepackage[utf8]{inputenc}

\usepackage{microtype}
\usepackage{makecell}

\usepackage{inconsolata}

\usepackage{graphicx}
\usepackage{times,latexsym}
\usepackage{url}
\usepackage[T1]{fontenc}
\usepackage{graphicx}
\usepackage{amsmath}
\usepackage{booktabs}
\usepackage{multirow} 
\usepackage[size=tiny]{todonotes}
\usepackage{subcaption}
\usepackage{makecell}
\usepackage{enumitem}
\usepackage{CJKutf8}
\newcommand{\naf}[1]{\textcolor{orange}{#1}}

\newcommand{\se}[1]{\textcolor{red}{#1}}
\newcommand{\cl}[1]{\textcolor{blue}{#1}}
\newcommand{\dl}[1]{\textcolor{teal}{#1}}

\title{
ContrastScore: Towards Higher Quality, Less Biased, More Efficient Evaluation Metrics with Contrastive Evaluation 

}

  \author{
  Xiao Wang\textsuperscript{1},
  Daniil Larionov\textsuperscript{2},
  Siwei Wu\textsuperscript{1},
  Yiqi Liu\textsuperscript{1},
  \\
  \textbf{Steffen Eger}\textsuperscript{3},
  \textbf{Nafise Sadat Moosavi}\textsuperscript{4},
  \textbf{Chenghua Lin}\textsuperscript{1}
  \\
  \textsuperscript{1}The University of Manchester, \textsuperscript{2}University of Mannheim,\\\textsuperscript{3}University of Technology Nuremberg, \textsuperscript{4}The University of Sheffield \\\texttt{\{xiao.wang-26,siwei.wu-2,yiqi.liu-6\}@postgrad.manchester.ac.uk}\\
  \texttt{daniil.larionov@uni-mannheim.de, steffen.eger@utn.de}\\\texttt{n.s.moosavi@sheffield.ac.uk, chenghua.lin@manchester.ac.uk }
}

\begin{document}

\maketitle

\begin{abstract}
%\naf{
Recent advances in automatic evaluation of natural language generation have increasingly relied on large language models as general-purpose metrics. While effective, these approaches often require high-capacity models, which introduce substantial computational costs, and remain susceptible to known evaluation pathologies, such as over-reliance on likelihood. We introduce ContrastScore, a contrastive evaluation paradigm that builds on the widely used BARTScore formulation by comparing token-level probabilities between a stronger and a weaker model. Instead of relying on single-model likelihoods or prompt-based judgments, ContrastScore captures disagreement between models to better reflect confidence and uncertainty in generation quality.
Empirical results on summarization and machine translation benchmarks show that ContrastScore, instantiated with paired moderate-scale models across both Qwen and LLaMA families, consistently outperforms larger alternatives, such as Qwen 7B and LLaMA 8B, in correlation with human ratings. In addition to improving evaluation quality, ContrastScore significantly reduces susceptibility to likelihood bias, offering a more robust and cost-effective alternative to larger LLM-based evaluation methods. \footnote{Source code is available at \url{https://github.com/sandywangxiao/ContrastScore}.}
% \footnote{Our code and evaluation scripts will be made publicly available upon publication.}
%}

%Evaluating the quality of generated text automatically remains a significant challenge. Conventional reference-based metrics have been shown to exhibit relatively weak correlation with human evaluations. Recent research advocates the use of large language models (LLMs) as source-based metrics for natural language generation (NLG) assessment. While promising, LLM-based metrics, particularly those using smaller models, still fall short in aligning with human judgments.
%In this work, we introduce ContrastScore, a contrastive evaluation metric designed to enable higher-quality, less biased, and more efficient assessment of generated text. We evaluate ContrastScore on two NLG tasks: machine translation and summarization. Experimental results show that ContrastScore consistently achieves stronger correlation with human judgments than both single-model and ensemble-based baselines. Notably, ContrastScore based on Qwen 3B and 0.5B even outperforms Qwen 7B, despite having only half as many parameters, demonstrating its efficiency. Furthermore, it effectively mitigates common evaluation biases such as length and likelihood preferences, resulting in more robust automatic evaluation. 
\end{abstract}

\section{Introduction}
Evaluating the quality of automatically generated text remains a fundamental challenge in natural language processing (NLP) and, in some cases, is nearly as difficult as generating the text itself. Traditional evaluation methods primarily rely on reference-based metrics such as BLEU, ROUGE, and METEOR ~\cite{papineni2002bleu,lin2004rouge,banerjee2005meteor,shen2023dense,pan-etal-2024-g,kalyan2024survey}, which are inadequate as they compare generated text to human-written references based on surface-level lexical overlap, often failing to capture semantic adequacy and fluency. Embedding-based metrics like BERTScore \cite{zhangbertscore} improve upon lexical approaches by leveraging contextualized representations for better semantic alignment. However, they remain sensitive to domain shifts, depend on high-quality references, and exhibit relatively weak correlations with human judgments \cite{zhao-etal-2023-evaluating}.  
To address these limitations, recent research has shifted towards source-based evaluation methods, %\todo{SE: not sure why the distinction between ref-free and ref-based is important for us. I also don't think ref-free eval is `safer' than ref-based}
leveraging large language models (LLMs). Metrics such as BARTScore \citep{yuan2021bartscore} and GPTScore \citep{fu-etal-2024-gptscore} assess generation quality using a single model’s probability distribution, while prompt-based LLM evaluators process structured prompts to evaluate text based on predefined criteria~\cite{que2024hellobench} or conduct comparative judgments without an explicit rubric~\cite{liusie2024llm}. Although these methods show promise, their effectiveness is inherently constrained by the underlying model’s capacity. Larger models generally perform better but are computationally expensive and incur high API costs \citep{larionov2024promptoptmeerrorawarepromptcompression}. Conversely, smaller models often have reduced capacity, resulting in unreliable evaluations. Furthermore, existing LLM-based metrics are susceptible to biases \citep{deutsch-etal-2022-limitations,sun-etal-2022-bertscore,lu-etal-2023-toward, liu-etal-2024-llms-narcissistic,hong2025beyond}, 
including over-reliance on likelihood, which may not align with human evaluation criteria. These challenges highlight the need for novel evaluation methods that are both efficient and reliable, balancing model capacity, accessibility, and evaluation robustness \citep{chen-eger-2023-menli,zhao-etal-2024-slide}.

Introduced by \citet{li-etal-2023-contrastive}, Contrastive Decoding enhances both diversity and factual accuracy by leveraging the disparity between a stronger \emph{expert model} and a weaker \emph{amateur model}, prioritizing tokens with the largest probability gap. Beyond decoding, contrastive principles have also been applied to model architectures, such as the Self-Contrast Mixture-of-Experts (SCMoE), where different expert pathways within a model act as internal contrastive filters to enhance reasoning \cite{NEURIPS2024_f77012b4}. These approaches highlight the potential of contrastive mechanisms in improving model outputs by using weaker models to generate contrastive signals, enabling stronger models to refine their predictions and achieve a balance between fluency and diversity.

Inspired by contrastive principles, we introduce \textbf{ContrastScore}, an evaluation metric that leverages structured disagreement between two models of differing capacities. Unlike conventional LLM-based evaluation methods that rely solely on a single model’s likelihood estimates, ContrastScore incorporates a weaker auxiliary model as a contrastive signal, dynamically adjusting probability scores to improve alignment with human judgements. 
Specifically, ContrastScore is built on a discrepancy-based probability formulation that measures the absolute difference between the probabilities assigned by two models. By utilizing discrepancies between a stronger (expert) model and a weaker (amateur) model, it generates more calibrated and robust evaluation scores. 
% \todo{SE: removed this paragraph. Our approach differs fundamentally from prior work in contrastive decoding \cite{li-etal-2023-contrastive}, which employs a log-probability ratio, comparing probabilities multiplicatively. While effective for generation tasks, this ratio-based formulation exponentially amplifies probability differences when the weaker model assigns low probabilities, leading to instability 
% %\todo{SE: do we show this empirically? If so, I would point it out here; otherwise it may sound as if this is common knowledge} 
% --- an undesirable property for evaluation metrics, where stability and calibration are crucial.
% In contrast, our proposed additive formulation ensures that probability differences remain bounded and interpretable, mitigating biases inherent in single-model evaluations. By avoiding extreme probability adjustments, ContrastScore provides a smoother and more stable correction mechanism, resulting in more reliable evaluation scores while maintaining computational efficiency.}
We conduct extensive experiments on two widely employed text generation tasks, namely, machine translation and summarization, to evaluate the effectiveness of ContrastScore. Our evaluation compares ContrastScore against established metrics and various baseline models, including single-model and ensemble-based approaches. Experimental results show that ContrastScore achieves a higher correlation with human judgments, outperforming both single-model and ensemble-based methods. Notably, ContrastScore using smaller models (Qwen 3B, Qwen 0.5B) even surpasses Qwen7B with 7.0\% of improvement in summarization %task 
despite having only half the parameters, demonstrating its efficiency. Furthermore, ContrastScore based on smaller models %significantly 
substantially 
enhances evaluation speed, providing at least a 1.5-fold increase in processing speed compared to a single larger model across both the Qwen and LLaMA families.  Additionally, it effectively mitigates the likelihood biase, enhancing robustness in automatic evaluation. By incorporating contrastive principles into evaluation, ContrastScore paves the way for a new paradigm of more robust and efficient text evaluation. 

The contributions of our paper are four-fold: 
\begin{itemize}[topsep=0pt,itemsep=-1ex,partopsep=1ex,parsep=1ex]
    \item We propose a simple yet highly effective difference-based formulation for contrastive evaluation, leveraging structured model discrepancies to produce more calibrated and reliable evaluation scores.
    \item We conduct extensive experiments across multiple generation tasks, diverse datasets, and various model families to rigorously assess the effectiveness of our approach.
    \item Contrastive evaluation strongly correlates with human judgments, outperforming single-model and ensemble methods
    %existing source-based metrics 
    while effectively addressing likelihood bias, which are prevalent in automatic evaluation.
    \item ContrastScore achieves significantly faster evaluation compared to larger single-model approaches, improving inference speed by at least 1.5 times while using only half the parameters, while maintaining comparable or even superior performance. 
    
\end{itemize}

\section{Related Work}
% \subsection{Evaluation Metrics for Text Generation}
%\paragraph{Reference-based Evaluation Metrics}

%Before the emergence of language \xw{models}, research  \xw{on evaluating} text generation mainly focused on simple 
%tasks with relatively standard answers (i.e., machine translation, and text summarization).
%\xw{metrics comparing} the n-gram overlap 
%metrics compare the overlap 
%between the model-generated text and the reference text, such as %F-score~\cite{melamed2003precision},

\paragraph{Automatic Evaluation of Text Generation.}
%\naf{
Automatic evaluation metrics for text generation can broadly be categorized into \textit{task-specific} and \textit{general-purpose} approaches. Early task-specific metrics such as BLEU~\cite{papineni2002bleu}, ROUGE~\cite{lin2004rouge}, and CHRF~\cite{popovic-2015-chrf} rely on surface-level n-gram overlap with a reference. While originally designed for tasks like machine translation and summarization, these metrics often fail to reflect actual semantic similarity due to the variability of valid natural language expressions.
To overcome these limitations, embedding-based metrics, such as BERTScore~\cite{zhangbertscore} and MoverScore~\cite{zhao-etal-2019-moverscore}, were introduced to compute token-level or document-level similarity using contextualized embeddings
% , achieving stronger correlation with human evaluations
. Further improvements came from task-specific metrics fine-tuned on human ratings, such as COMET~\cite{rei-etal-2020-comet}, COMET-KIWI~\cite{rei2022cometkiwi}, 
% XCOMET~\cite{guerreiro-etal-2024-xcomet}, 
BLEURT~\cite{sellam-etal-2020-bleurt}, and Prism~\cite{thompson-post-2020-automatic}. These models learn to directly predict human preferences but often generalize poorly across domains and tasks.
General-purpose evaluation has increasingly shifted toward leveraging LLMs as evaluators. Prompt-based approaches such as G-Eval~\cite{liu-etal-2023-g}, ChatEval~\cite{chan2024chateval,hong2025beyond}, and GPT-based judge systems treat LLMs as reference-free annotators that directly assess text quality via carefully crafted prompts. While these methods often achieve high correlation with human judgments, they typically rely on proprietary models like GPT-4, making them costly, 
% difficult to reproduce, 
and sensitive to prompt phrasing~\cite{leiter-eger-2024-prexme}.
An alternative line of work explores probability-based evaluation metrics, including BARTScore~\cite{yuan2021bartscore} and GPTScore~\cite{fu-etal-2024-gptscore}.
% , and MAUVE~\cite{pillutla2021mauve}. 
BARTScore and GPTScore share a common formulation: they compute the conditional log-likelihood of a generation given context, using a pretrained language model as the scoring function. 
% \textcolor{red}{The performance of these methods depends heavily on the choice of the underlying model.} 
% MAUVE, while also probabilistic, instead measures divergence between the overall distributions of human and model-generated text, making it more suitable for analyzing global generation behavior than conditional quality.
% BARTScore has seen especially broad adoption due to its simplicity, reproducibility, and flexibility. It can be implemented using publicly available encoder-decoder models (e.g., BART, T5), and achieves moderate correlation with human ratings across summarization, translation, and dialogue tasks, even without access to proprietary LLMs.
However, these evaluation methods inherit the biases and limitations of the underlying model \cite{ohi-etal-2024-likelihood}. 

\paragraph{Contrastive Paradigm.} 
Contrastive methods have recently gained attention in text generation, where the differences between models of varying capacities are leveraged to improve output quality \cite{su2022contrastive, o2023contrastive,li-etal-2023-contrastive}. Rather than relying on ensemble methods or majority voting, contrastive decoding explicitly uses a weaker model as a contrastive signal to refine or steer the output of a stronger model. This approach has demonstrated success in mitigating common generation issues such as repetition, overconfidence, and incoherence.
% \citet{li-etal-2023-contrastive} utilizes a large `\textit{expert}' model and a smaller `\textit{amateur}' model. Instead of directly following the expert’s probabilities, contrastive decoding selects tokens that maximize the log-probability difference between the two models. This results in text that is more fluent, diverse, and factually accurate.
Moreover, contrastive principles have been extended to model architecture optimization. \citet{NEURIPS2024_f77012b4} propose the Self-Contrast Mixture-of-Experts (SCMoE) framework, where different expert pathways within the same model act as internal contrastive filters, which improves reasoning capabilities and overall model performance.
%, eliminating the need for external weaker models. 
Note that contrastive decoding is also related to language model arithmetic in which arithmetic combinations of LLMs are considered for adjusted generation \citep{dekoninck2024controlled}. This includes subtraction, which leads to output favored by one LLM but disfavored by the other, which could be leveraged in adversarial settings \citep{Zhang2024LLMbasedMP}.

Inspired by these advancements, as described in \S\ref{sec:contrastscore}, we extend contrastive principles beyond generation to the domain of evaluation. Instead of refining model outputs, we propose %Contrastive Evaluation
ContrastScore---a novel metric with a %fundamentally 
different formulation from prior contrastive decoding work. 

\section{Methodology}

% The primary goal of this study is to introduce a contrastive evaluation framework,
% %that extends BARTScore into a new metric, 
% \textit{ContrastScore}. %Instead of relying on a single model to assess the quality of the generated text, ContrastScore leverages the probabilities from both an expert model and an amateur model, enabling a more discriminative and robust assessment of text quality.
% Current generative metrics for text evaluation are 
% predominantly
% based on a single language model.  %\citep{yuan2021bartscore, qin-etal-2023-t5score, fu-etal-2024-gptscore}. 
% However, their performance is highly dependent on the capabilities of the underlying model, leading to performance disparities, where metrics based on smaller or medium-sized models typically underperform those using larger models. Additionally, these single-model approaches are inherently affected by the biases of the underlying model such as likelihood bias, where the model may over-prefer fluent but less accurate outputs, or pretrained biases that can influence evaluation outcomes in unintended ways.

% To address this, we introduce \emph{ContrastScore}, a method that leverages the structured disagreement between two models of different capacities to produce a bias-aware, dynamically calibrated probability score for evaluation. Rather than relying solely on the probability of a single model, ContrastScore incorporates an auxiliary model as a contrastive signal, allowing for a more refined probability adjustment that improves alignment with human judgments.

The primary goal of this study is to overcome limitations of single-model evaluators, such as capacity bias and over-reliance on model-specific likelihoods, by capturing where strong and weak models disagree in token-level probabilities. This section first reviews generative likelihood-based evaluation and contrastive decoding, then presents the design and formulation of ContrastScore.

%\subsection{Problem formulation}

%As previously mentioned, the primary objective of this study is to evaluate the quality of the generated text without references. Specifically, we concentrate on conditional text generation tasks (e.g., summarization), where the aim is to generate a hypothesis $\textbf{h}\text{=}(h_1,h_2,...,h_m)$ from a given source text $\textbf{s}\text{=}(s_1,s_2,...,s_n)$. Our goal is to find methods to evaluate $\textbf{h}$:

%\begin{equation}
%\text{Evaluator}(\textbf{h}) = f(\textbf{s},\textbf{h},\theta)
%\end{equation}
%where $\theta$ is the parameters.

%\subsection{BARTScore} 
% As source-based evaluation, generative evluation metrics are widely used for assesssing the generated text. 

%Recent studies have explored the use of large language models as source-based metrics for natural language generation (NLG) assessment. BARTScore \cite{yuan2021bartscore} introduced the generative evaluation metrics, conceptualizing text assessment as a generation problem. It evaluates the quality of the generated text by leveraging the probability of the text under a pretrained language model. While it was initially designed for the BART model \cite{lewis-etal-2020-bart},
%it can be adapted to any language model(e.g. GPTScore\cite{fu-etal-2024-gptscore}, T5Score\cite{qin-etal-2023-t5score}). We generalize this generative evaluation method based on any single generation model as follows:

\subsection{Generative Evaluation}
Our evaluation framework builds on probability-based text assessment methods, such as BARTScore, which estimate quality through log-likelihood computations under a pretrained generative model. Given a generated hypothesis $\textbf{h}$, its quality is evaluated as:
\begin{equation} 
\text{Score}(\textbf{h} | d, \mathcal{S}) = \sum_{t=1}^{m}w_t \log P(h_t | \textbf{h}_{<t}, \mathcal{S}, \theta)
\end{equation}
where $\mathcal{S}$ represents the supplementary text, which may consist of the source text $\textbf{s}$ (in a source-based setting) or the reference output $\textbf{r}$ (in a reference-based setting). $d$ denotes the evaluation dimension (e.g., fluency), $w_t$ \footnote{Following prior studies \cite{yuan2021bartscore, qin-etal-2023-t5score}, $w_t$ is weighed equally for all tokens in this work.} refers to the token-level weight, and $\theta$ is the model parameters. %$w_t$ refers to the weight assigned to each token $h_t$, which is typically weighed equally for all tokens in prior studies \cite{yuan2021bartscore,qin-etal-2023-t5score}.

\subsection{Contrastive Decoding}

% Assume an expert and an amateur model that assign probabilities 
% $p_{\text{EXP}}^t = p(h_t \mid h_{<t}, \mathcal{S}, \theta_{\text{EXP}})$ and 
% $p_{\text{AMA}}^t = p(h_t \mid h_{<t}, \mathcal{S}, \theta_{\text{AMA}})$ 
% to the next token $h_t$, where $\theta_{\text{EXP}}$ and $\theta_{\text{AMA}}$ 
% denote the parameters of the expert and amateur models, respectively.
%Assume an expert and amateur model assigning probabilities $p_{\text{EXP}}^{t} = p(h_t \mid h_{<t}, \mathcal{S}, \theta_{\text{EXP}}) $ and $p_{\text{AMA}}^{t} = p(h_t \mid h_{<t}, \mathcal{S}, \theta_{\text{AMA}})$ to the next token $h_t$, respectively.

Assume an expert and amateur model assigning probabilities
\begin{equation}
   p_{\text{EXP}}^{t} = p(h_t \mid h_{<t}, \mathcal{S}, \theta_{\text{EXP}})
\end{equation}
\begin{equation}
   p_{\text{AMA}}^{t} = p(h_t \mid h_{<t}, \mathcal{S}, \theta_{\text{AMA}})
\label{eq:ama_prob}
\end{equation}
to the next token $h_t$, where 
$p_{\text{EXP}}^{t}$ and $p_{\text{AMA}}^{t}$ represent the probabilities %assigned by 
of the expert and amateur models, respectively, %to the generated token $h_t$. 
for $h_t$ and where  
$\theta_{\text{EXP}}$ and $\theta_{\text{AMA}}$ refer to the parameters of the expert and amateur model. \citet{li-etal-2023-contrastive} propose the \emph{contrastive decoding} objective  
\begin{align}\label{eq:cdscore}
\text{CD-Score} = \begin{cases}
\log{\frac{p_{\text{EXP}}^{t}}{p_{\text{AMA}}^{t}}} & \text{if } h_t\in V_{\text{head}}(h_{<t}) \\
-\infty & \text{else}
\end{cases}
\end{align}
% \begin{align}\label{eq:cdscore}
% = \begin{cases}
% \log{p_{\text{EXP}}^{t}}-\log{p_{\text{AMA}}^{t}} & \text{if } h_t\in V_{\text{head}}(h_{<t}) \\
% -\infty & \text{else}
% \end{cases}
% \end{align}
Here, $V_{\text{head}}(h_{<t})$ selects the top most likely tokens $x_t$ under the expert model given history $h_{<t}$. During next token prediction, CD-Score effectively only considers the most likely tokens of the expert model as candidates, and then assigns modified logits %probabilities 
%\log{\frac{p_{\text{EXP}}^{t}}{p_{\text{AMA}}^{t}}}$
$\log{\frac{p_{\text{EXP}}^{t}}{p_{\text{AMA}}^{t}}}$
% =\log{p_{\text{EXP}}^{t}}-\log{p_{\text{AMA}}^{t}}$ 
to them. The intuition behind CD-Score is to consider the most likely expert tokens as continuation for generation but then choose those tokens among them which the amateur might not favor (e.g., repetitive tokens). 

%\textcolor{blue}{Unfortunately, CD-Score is not well-suited for evaluation tasks because its ratio-based formulation is designed for generation tasks, where it leverages relative differences in token probabilities across models to select the most appropriate token. In contrast, evaluation tasks treat probabilities as measures of confidence in each generated token, emphasizing the overall scale rather than relative differences. As a result, the ratio-based approach is unable to distinguish scenarios where both expert and amateur models assign similarly high or low probabilities. For example, if the two models both assign probabilities of 0.9 or 0.1, CD-Score remains the same and thus fails to capture any meaningful difference in absolute confidence.}

Unfortunately, CD-Score is unsuitable for evaluation because every generated sequence $\textbf{h}$ containing a token $h_t$ not in $V_{\text{head}}(h_{<t})$ would receive a score of $-\infty$, essentially being ruled out as a candidate, even though it should have been considered as part of the sequence under evaluation. 
%(in practice one might exchange $-\infty$ with a very low negative value, mitigating the issue to a certain degree). 
As an alternative, one might consider the division score $\log{\frac{p_{\text{EXP}}^{t}}{p_{\text{AMA}}^{t}}}$ %=\log{p_{\text{EXP}}^{t}}-\log{p_{\text{AMA}}^{t}}$ 
% \begin{equation}\label{eq:division}
%     \log{\frac{p_{\text{EXP}}^{t}}{p_{\text{AMA}}^{t}}} =\log{p_{\text{EXP}}^{t}}-\log{p_{\text{AMA}}^{t}}
% \end{equation}
itself as an objective for evaluation. However, as pointed out by \citet{li-etal-2023-contrastive}, this approach suffers from penalizing many standard text sequences as unlikely, for which both expert and amateur would hold the same probabilities. For example, if both models assign a probability of 0.9 or 0.1 to a token, the score remains the same, making it unable to distinguish between confidently correct and incorrect predictions. As a result, this approach fails to highlight important distinctions between tokens that should matter in evaluation. The division score also becomes instable especially when the amateur probability is close to zero.%, distorting the evaluation score.

\subsection{ContrastScore} \label{sec:contrastscore} 
%Given a larger (expert) model and a smaller (amateur) model,  ContrastScore leverages the divergence between two single models with a discrepancy-aware probability:
To address this issue \textit{for evaluation}, we propose a subtraction-based contrastive formulation with a scaling factor $\gamma \in [0,1]$ that downweights the amateur model. In this way, we naturally retain all tokens in the input sequence for evaluation, as well as a property that favors generations preferred by the expert model, similar to CD-Score. Importantly, the subtraction-based formulation ensures that the scores of individual tokens remain comparable across positions---unlike the division-based approach, where small variations in the amateur model's likelihood can lead to disproportionately large and unstable scores. By using a small $\gamma$, we control the influence of the amateur model’s probability and mitigate such instability, yielding more reliable evaluation signals. Empirically, we choose a formula for contrastive evaluation that leverages the absolute value distance between the expert and the scaled down amateur:

%\se{To address this issue \emph{for evaluation}, we propose a scaling factor $\gamma \in [0,1]$ \textcolor{red}{downweighting} the amateur model. In this way, we have a contrastive signal as in CD-Score 
% which prefers generations of the expert model 
% but %one that does not rule out all plausible standard text sequences indiscriminately as Eq.~\eqref{eq:division} would do. On the other hand, in this way, 
% we %also 
% do not run into the issue of assigning probability of $-\infty$ to sequences that contain unlikely expert tokens, as Eq.~\eqref{eq:cdscore} would cause. On the other hand, we also do not rule out all plausible standard text sequences indiscriminately as Eq.~\eqref{eq:division} would do.}

%\footnote{Empirically, one might consider alternative formulations, including $\log{\frac{p_{\text{EXP}}^{t}}{\gamma p_{\text{AMA}}^{t}}} =\log{p_{\text{EXP}}^{t}}-\gamma\log{p_{\text{AMA}}^{t}}$. In our experiments, these were less effective, however.}
% \begin{equation}
%   \text{ContrastScore} = \sum_{t=1}^{m} %w_t 
%   \log \left( \left| p_{\text{EXP}}^{t} - \gamma  p_{\text{AMA}}^{t} \right| \right)
% \label{eq:contrastscore}
% \end{equation}
\vspace{-8pt}
\begin{equation}
  \text{ContrastScore} = \sum_{t=1}^{m} w_t 
  \log \left( \left| p_{\text{EXP}}^{t} - \gamma  p_{\text{AMA}}^{t} \right| \right) 
\label{eq:contrastscore}
\end{equation}

At each time step $t$, ContrastScore rewards tokens for which expert and (downweighted) amateur generation probabilities are maximally distinct (as the absolute value is a distance function between two distributions). This would ensure that ContrastScore leverages the strengths of the expert model but removes the limitations of the amateur model; %in practice, it might favor (for example) generations which are less repetitive to which the amateur model inclines. 
in practice, it might for example lean toward generations that are less repetitive, which the amateur model tends %to produce.
to 
favor. 
Note, however, that there are probability ranges where the scaled down amateur would be preferred over the expert---e.g., %with $\gamma=1$, ContrastScore would be indifferent between the expert and the amateur but indiscriminately favor disagreement between them.}
when $p^t_{\text{EXP}}=0$, then our formula would still assign the scaled probability of the amateur to $h_t$. With $\gamma=1$, ContrastScore would be indifferent between the expert and the amateur but indiscriminately favor disagreement between them, an undesirable behavior similar to the division score above.

\section{Experimental Setup}
\label{sec:experiments}
\iffalse
We evaluate the performance and robustness of the proposed ContrastScore on two tasks: machine translation and summarization, using models from the LLaMA and Qwen families to ensure generalizability.  
To benchmark ContrastScore, we compare it against single-model evaluators as well as an ensemble setting, where the probabilities of the expert and amateur models are averaged per token. The ensemble probability at token $t$, denoted as $p_{\text{Ens}}^t$, is computed as: $
p_{\text{Ens}}^t = \frac{1}{2} \left( p_{\text{EXP}}^t + p_{\text{AMA}}^t \right)$.
\fi

\subsection{Datasets and Tasks}

For summarization, we use the \textbf{SummEval} dataset~\cite{fabbri2021summeval}, which contains model-generated summaries of CNN/DailyMail articles, as well as \textbf{QAGS-XSUM}~\cite{wang2020asking}, which includes 239 system outputs on the XSUM dataset. For machine translation, we use the \textbf{MQM22} and \textbf{MQM23} datasets from the WMT22~\cite{freitag2022results} and WMT23~\cite{freitag2023results} Metrics Shared Tasks. The language pairs involved represent a diverse range of translation scenarios, including high-resource (EN-DE), typologically distant (ZH-EN and EN-RU), and low-resource (HE-EN) settings, thereby enabling a comprehensive evaluation of our approach. See Appendix\ref{appen:dataset} for more details.

\subsection{Baseline Metrics}
We consider the following baseline metrics for comparison:
\textbf{BLEU}~\cite{papineni2002bleu}, \textbf{CHRF}~\cite{papineni2002bleu}, \textbf{ROUGE 1}, \textbf{ROUGE 2}, \textbf{ROUGE L}~\cite{lin2004rouge}, \textbf{COMET}~\cite{rei-etal-2020-comet},~\textbf{COMET-KIWI}~\cite{rei2022cometkiwi}, 
% \textbf{XCOMET}~\cite{guerreiro-etal-2024-xcomet}, 
\textbf{BERTScore}~\cite{zhangbertscore}, \textbf{MoverScore}~\cite{zhao-etal-2019-moverscore}, and \textbf{BARTScore}~\cite{yuan2021bartscore}. 
% \todo{SE: I think we MUST include some recent SOTA metric for comparison, e.g., XCOMET-XXL, MetricX-24, etc.}
The details of these metrics can be found in Appendix \ref{appen:metrics}. In addition, we compare it against \textbf{Single-model} evaluators as well as an \textbf{Ensemble} setting, where the probabilities of the expert and amateur models are averaged per token, using models from the LLaMA and Qwen families to ensure generalizability. The ensemble probability at token $t$, denoted as $p_{\text{Ens}}^t$, is computed as: 
%$p_{\text{Ens}}^t = \frac{1}{2} \left( p_{\text{EXP}}^t + p_{\text{AMA}}^t \right)$.
\begin{equation}
 p_{\text{Ens}}^t = \frac{1}{2} \left( p_{\text{EXP}}^t + p_{\text{AMA}}^t \right)   
\end{equation}

\subsection{Meta-Evaluation}
\label{sec:meta-evalation}

We assess the evaluation metrics in terms of quality, biases, and efficiency.
\vspace{5pt}

\noindent\textbf{Correlation with Human Scores (Quality).}~~The effectiveness of an evaluator is measured using the Pearson correlation between its scores and human scores, where a higher correlation indicates stronger alignment with human judgments.  
For translation, Pearson correlation assesses the evaluator's ability to capture differences between candidate translations based on the source texts, comparing them to human judgment scores derived from fine-grained MQM annotations~\citep{freitag-etal-2021-experts}.  
For summarization, we compute Pearson correlation with human judgments across different quality aspects: coherence, fluency, consistency, relevance, and factuality.

%The task involves computing quality scores for candidate summaries with respect to the source texts. We then measure Pearson correlation with human judgments across four quality aspects: coherence, fluency, consistency, and relevance.

\vspace{5pt}

\noindent\textbf{Bias Evaluation.}~~We measure biases in likelihood bias.
% \noindent\textit{\underline{Likelihood Bias}}: 
%LLMs generate predictions by estimating the likelihood of text based on patterns learned during training, optimizing for high-probability outputs. 
Recent studies have shown that LLMs tend to favor sentences they deem more likely, often assigning higher evaluation scores and performing better on such sentences, regardless of the specific task or evaluation criteria \cite{ohi-etal-2024-likelihood,doi:10.1073/pnas.2322420121,mccoy2024language}.
\citet{ohi-etal-2024-likelihood} find that LLM-based evaluators systematically overrate high-likelihood sentences and underrate low-likelihood ones compared to human scores. 
To quantify this likelihood bias, they propose BiasScore
defined as:
\begin{equation}
\text{BiasScore} = \rho(\textit{LS}, \textit{US})
\label{eq:prob_bias}
\end{equation}

where $\rho$ is the Spearman correlation coefficient,
$\textit{LS}$ (Likelihood Score) represents the probability $P$ assigned by the LLM, and $\textit{US}$ (Unfairness Score) captures the discrepancy between evaluator scores and human scores.
%Likelihood bias assesses the influence of base model on the evaluation metric. 
Likelihood bias measures the extent to which an evaluation metric is influenced by the model’s inherent probability estimates. A \textit{lower BiasScore} indicates that the metric is less dependent on the model's likelihood biases.%, leading to more human-aligned evaluations.
%A lower likelihood bias value indicates that the evaluation metric is less dependent on the inherent likelihood of the specific model.

\vspace{5pt}
\noindent\textbf{Efficiency.}~~It is measured by the number of samples processed per second, with a higher processing speed indicating greater efficiency.

\subsection{Hyperparameters and Environment}
\label{hyparameter}
%To ensure robust generalization, 
We use models from the same family but of different sizes, where a larger model serves as the expert and a smaller model as the amateur. Specifically,
%To demonstrate our method's generalization across model families and sizes, 
we use LLaMA3.2-Instruct (1B, 3B) and LLaMA3.1-Instruct (8B), as the LLaMA 3.2 version does not include an 8B model. Additionally, we evaluate Qwen2.5-Instruct (0.5B, 3B, and 7B).\footnote{See the discussion in the Limitations section on why we use models from the same family.% for both expert and amateur models. 
%We use models of the same family for expert and amateur to avoid inconsistencies in token splits caused by different tokenizers across model families.
% We use models of the same family for expert and amateur, as models from different families use different tokenization.
}
%We also attempted to explore ContrastScore using models from different families; however, this proved problematic due to differences in model tokenization. 
%\todo{SE: perhaps in the limitations, we should mention that using models from different families is problematic because of tokenization}
In all our experiments\footnote{
%We set the decoding temperatures for both the expert and amateur models, denoted as $\tau_{1}$ and $\tau_{2}$, respectively. 
Following \citet{li-etal-2023-contrastive}, we set the decoding temperatures for the expert and amateur to 0.5 and 1.5, respectively.
%$\tau_{2} = 0.5$, while $\tau_{1}$ is set to 1.5.
}, the scaling factor for the amateur model, $\gamma$, is set to 0.1.\footnote{This setting is based on our pilot investigation, which found that $\gamma = 0.1$ gives the best overall performance empirically 
for ZH-EN in MT (Figure \ref{gamma} in Appendix).} 
All experiments were run on a GPU cluster under same conditions and node configuration to ensure fair comparisons. Each node of the cluster has 4$\times$H100 GPUs.

\section{Results}\label{sec:results}

We evaluate ContrastScore in terms of its correlation with human scores, bias, and efficiency, compared to single-model and ensemble methods.

\subsection{Correlation with Human Scores}

\newcommand{\ul}[1]{\underline{#1}}
\newcommand{\bt}[1]{\textbf{#1}}

\begin{table*}[!ht]
\footnotesize
    \centering
    \scalebox{0.95}{
    \begin{tabular}{lllcccccc}
        \toprule
        \multirow{2}{*}{}&\multirow{2}{*}{}&\multirow{2}{*}{Evaluators}& \multicolumn{1}{l}{QAGS-XSUM} & \multicolumn{4}{c}{SummEval} & \multirow{2}{*}{AVG} \\
        \cmidrule(lr){4-4} \cmidrule(lr){5-8}
        && & Factuality & Coherence & Consistency & Fluency & Relevance & \\
        \midrule
        \multicolumn{2}{c}{\multirow{6}{*}{Baseline}} 
         & ROUGE-1      & 0.055     & 0.238     & 0.079     & 0.071     & 0.330     & \cellcolor{blue!10}0.154 \\
        && ROUGE-2      & \ul{0.121} & 0.164     & 0.073     & 0.066     & 0.229     & \cellcolor{blue!10}0.130 \\
        && ROUGE-L      & 0.075     & 0.207     & 0.071     & 0.078     & 0.303     & \cellcolor{blue!10}0.147 \\
        && MoverScore   & 0.028     & 0.129     & 0.141     & 0.133     & 0.280     & \cellcolor{blue!10}0.142 \\
        && BERTScore    & 0.024     & 0.342     & 0.115     & 0.156     & 0.372     & \cellcolor{blue!10}0.202 \\
        && BARTScore    & 0.099     & \ul{0.450} & \ul{0.339} & \ul{0.343} & \bt{0.428} & \cellcolor{blue!10}\ul{0.332} \\
        \hline \hline
        \multirow{10}{*}{LLaMA} & \multirow{3}{*}{Single} 
          & 1B          & 0.064     & 0.417     & 0.549     & 0.542     & 0.344     & \cellcolor{blue!10}0.383 \\
        & & 3B          & 0.157     & 0.388     & 0.575     & 0.530     & 0.338     & \cellcolor{blue!10}0.398 \\
        & & 8B          & 0.205     & 0.368     & \ul{0.588} & 0.539     & 0.332     & \cellcolor{blue!10}0.406 \\
        \cmidrule(l){2-9}
        & \multirow{3}{*}{Ensemble} 
          & (3B,1B)     & 0.170     & 0.427     & 0.564     & 0.542     & 0.344     & \cellcolor{blue!10}0.409 \\
        & & (8B,1B)     & 0.212     & 0.429     & 0.581     & 0.556     & 0.357     & \cellcolor{blue!10}0.427 \\
        & & (8B,3B)     & 0.241     & 0.414     & 0.587     & 0.544     & 0.351     & \cellcolor{blue!10}0.428 \\
        \cmidrule(l){2-9}
        & \multirow{3}{*}{Contrast} 
          & (3B,1B)     & 0.262     & 0.456     & 0.554     & 0.551     & 0.334     & \cellcolor{blue!30}0.431 \\
        & & (8B,1B)     & \bt{0.324} & 0.461     & 0.584     & \bt{0.575} & 0.368     & \cellcolor{blue!30}0.462 \\ 
        & & (8B,3B)     & 0.323     & \bt{0.462} & 0.587     & \bt{0.575}     & \ul{0.369} & \cellcolor{blue!30}\bt{0.463} \\
        \hline \hline
        \multirow{10}{*}{Qwen} & \multirow{3}{*}{Single} 
          & 0.5B        & 0.107     & 0.386     & 0.536     & 0.520     & 0.315     & \cellcolor{blue!10}0.373 \\
        & & 3B          & 0.112     & 0.370     & 0.544     & 0.492     & 0.319     & \cellcolor{blue!10}0.367 \\
        & & 7B          & 0.091     & 0.371     & 0.560     & 0.510     & 0.327     & \cellcolor{blue!10}0.372 \\
        \cmidrule(l){2-9}
        & \multirow{3}{*}{Ensemble} 
          & (3B,0.5B)   & 0.081     & 0.394     & 0.568     & 0.536     & 0.331     & \cellcolor{blue!10}0.382 \\
        & & (7B,0.5B)   & 0.095     & 0.386     & 0.577     & 0.538     & 0.333     & \cellcolor{blue!10}0.386 \\ 
        & & (7B,3B)     & 0.103     & 0.378     & 0.584     & 0.521     & 0.330     & \cellcolor{blue!10}0.383 \\
        \cmidrule(l){2-9}
        & \multirow{3}{*}{Contrast} 
          & (3B,0.5B)   & 0.128     & \ul{0.397} & 0.594     & 0.546     & 0.323     & \cellcolor{blue!30}0.398 \\
        & & (7B,0.5B)   & 0.122     & 0.376     & \bt{0.608} & 0.554     & 0.322     & \cellcolor{blue!30}0.396 \\
        & & (7B,3B)     & \ul{0.134} & 0.384     & 0.606     & \ul{0.557} & \ul{0.334} & \cellcolor{blue!30}\ul{0.403} \\
        \bottomrule
    \end{tabular}
    }
    \caption{Pearson correlation of evaluators with human scores in summarization.  \bt{boldface} represents best overall scores, while \ul{underline} represents best scores within each model group (Baseline, LLaMA, Qwen).  Overall, ContratScore outperforms single and ensemble methods as well as baseline metrics for both LLaMA and Qwen families. 
    % Scores where the contrast of expert and amateur outperforms both the single expert model and the ensemble method are highlighted in \colorbox{blue!30}{darker purple}.
   } 
    \label{tab:summ} 
\end{table*}

\noindent\textbf{Summarization.}~~Table \ref{tab:summ} presents the Pearson correlation between different evaluators and human scores for summarization. Our results indicate that ContrastScore provides best results overall. Furthermore, ContrastScore consistently surpasses single models while utilizing much fewer parameters. Specifically, ContrastScore with Qwen(3B, 0.5B) outperforms the single Qwen 7B by 7.0\%, and ContrastScore with LLaMA(3B, 1B) exceeds single LLaMA 8B by 6.2\%. 
These results reinforce the observation that ContrastScore can efficiently enhance model evaluation by leveraging smaller models.
Besides improving overall correlation with human scores, ContrastScore demonstrates notable gains across key summarization dimensions, particularly coherence, fluency, relevance and factuality. When applied to LLaMA(8B, 3B), it enhances coherence by 25.5\%, 
%0.094 25.54
fluency by 6.7\%, 
%0.036  6.68
relevance by 11.1\%,
%0.037  11.14
and factuality by 57.6\%, 
compared to the single LLaMA 8B. However, its impact on consistency differs across model families. 

\vspace{3pt}

\noindent\textbf{Machine Translation.}~~Table~\ref{tab:mt} presents the Pearson correlation between various evaluators and human scores.
The results show that, overall, ContrastScore outperforms both single-model and ensemble approaches.
For example, ContrastScore based on Qwen(7B, 3B) improves correlation by 
7.1\% over the single Qwen 7B model and by 4.6\% over the ensemble of Qwen(7B, 0.5B).
In contrast, ensembling does not always improve performance:  
While ensemble methods are generally expected to refine predictions, they can sometimes reduce correlation with human scores. For instance, ensembling LLaMA (8B, 1B) leads to a 3.3\% decrease in correlation compared to the LLaMA 8B model. Similar trends can be observed in pairwise accuracy \cite{deutsch-etal-2023-ties}, as shown in Table \ref{tab:pairAcc} of the Appendix.

\vspace{-1pt}

Beyond the cases when the amateur model is too weak to provide a meaningful correction signal---such as the low correlation scores observed in the HE-EN pair for both LLaMA 1B and 3B (0.32 and 0.29, respectively), which fail to offer reliable contrastive feedback against LLaMA 8B---our results demonstrate that ContrastScore can achieve correlation scores comparable to, or even exceeding, those of the largest model using only two smaller models. Specifically, ContratScore of Qwen(3B, 0.5B) surpasses single Qwen 7B by 3.1\%, improving from 0.449 to 0.463 with only half the parameters.

\begin{table*}[!ht]
    \centering
    \footnotesize
    \scalebox{0.95}{
    \begin{tabular}{lllccccccc}
        \toprule
        \multirow{2}{*}{}&\multirow{2}{*}{}&\multirow{2}{*}{Evaluators} & \multicolumn{3}{c}{MQM22} & \multicolumn{3}{c}{MQM23} & \multirow{2}{*}{AVG} \\
        \cmidrule(lr){4-6} \cmidrule(lr){7-9}
        & && EN-DE & ZH-EN & EN-RU & EN-DE & ZH-EN & HE-EN  & \\
        \midrule
        \multicolumn{2}{c}{\multirow{6}{*}{Baseline}} 
         & BLEU         & 0.194     & 0.156     & 0.142     & 0.163     & 0.085     & 0.208     & \cellcolor{blue!10}0.158 \\
        && CHRF         & 0.236     & 0.156     & 0.169     & 0.232     & 0.063     & 0.244     & \cellcolor{blue!10}0.183 \\
        && BERTScore    & 0.263     & 0.311     & 0.197     & 0.325     & 0.236     & 0.336     & \cellcolor{blue!10}0.278 \\
        && BARTScore    & 0.254     & 0.287     & 0.201     & 0.201     & 0.182     & 0.317     & \cellcolor{blue!10}0.240 \\
        && COMET        & \bt{0.476}& \ul{0.403}& \bt{0.417}& 0.432     & 0.396     & \ul{0.417}& \cellcolor{blue!10}\ul{0.424} \\
        && COMET-KIWI   & 0.392     & 0.367     & 0.354     & \ul{0.475}& \ul{0.442}& 0.395   & \cellcolor{blue!10}0.404\\
        % &&XCOMMET-XXL   & 0.724 & 0.678 & 0.744 & 0.660 & 0.597 & 0.482 & \cellcolor{blue!10}0.648     \\
        \hline \hline
        \multirow{10}{*}{LLaMA} & \multirow{3}{*}{Single} & 
            1B          & 0.255     & 0.371     & 0.286     & 0.530     & 0.518     & 0.320     & \cellcolor{blue!10}0.380 \\
        & & 3B          & 0.284     & 0.366     & 0.293     & 0.578     & 0.526     & 0.293     & \cellcolor{blue!10}0.390 \\
        & & 8B          & 0.363     & 0.376     & 0.356     & 0.632     & 0.574     & \bt{0.491}& \cellcolor{blue!10}0.465 \\
        \cmidrule(l){2-10}
        & \multirow{3}{*}{Ensemble} & 
            (3B,1B)     & 0.287     & 0.371     & 0.307     & 0.568     & 0.538     & 0.306     & \cellcolor{blue!10}0.396 \\
        & & (8B,1B)     & 0.325     & 0.382     & 0.354     & 0.605     & 0.575     & 0.462     & \cellcolor{blue!10}0.450 \\
        & & (8B,3B)     & 0.337     & 0.379     & 0.348     & 0.619     & 0.574     & 0.453     & \cellcolor{blue!10}0.452 \\ 
        \cmidrule(l){2-10}
        & \multirow{3}{*}{Contrast} & 
            (3B,1B)     & 0.338     & 0.382     & 0.347     & 0.599     & 0.562     & 0.284     & \cellcolor{blue!30}0.419 \\
        & & (8B,1B)     & \ul{0.392}& 0.391     & 0.406& \bt{0.641}& 0.590     & 0.484     & \cellcolor{blue!30}\bt{0.484} \\ 
        & & (8B,3B)     & 0.383     & \ul{0.393}& \ul{0.409}     & 0.639     & \ul{0.595}& 0.482     & \cellcolor{blue!30}0.483 \\ 
        \hline \hline
        \multirow{10}{*}{Qwen} & \multirow{3}{*}{Single} 
        & 0.5B          & 0.222     & 0.394     & 0.294     & 0.487     & 0.557     & 0.347     & \cellcolor{blue!10}0.383 \\
        & & 3B          & 0.306     & 0.413     & 0.299     & 0.573     & 0.594     & 0.415     & \cellcolor{blue!10}0.433 \\
        & & 7B          & 0.326     & 0.419     & 0.330     & 0.600     & 0.574     & 0.445     & \cellcolor{blue!10}0.449 \\
        \cmidrule(l){2-10} 
        & \multirow{3}{*}{Ensemble} 
          & (3B,0.5B)   & 0.290     & 0.408     & 0.315     & 0.548     & 0.599     & 0.421     & \cellcolor{blue!10}0.430 \\
        & & (7B,0.5B)   & 0.301     & 0.412     & 0.331     & 0.567     & 0.598     & 0.446     & \cellcolor{blue!10}0.443 \\
        & & (7B,3B)     & 0.333     & 0.424     & 0.341     & 0.599     & 0.608     & 0.456     & \cellcolor{blue!10}0.460 \\
        \cmidrule(l){2-10}
        & \multirow{3}{*}{Contrast} 
          & (3B,0.5B)   & 0.342     & 0.432     & 0.351     & 0.590     & 0.628     & 0.431     & \cellcolor{blue!30}0.463 \\
        & & (7B,0.5B)   & 0.359     & 0.435     & 0.382     & \ul{0.611}& 0.628     & 0.465     & \cellcolor{blue!30}0.480 \\
        & & (7B,3B)     & \ul{0.362}& \bt{0.439}& \ul{0.384}& 0.605     & \bt{0.629}& \ul{0.466}& \cellcolor{blue!30}\ul{0.481} \\ 
        \bottomrule
    \end{tabular}
    }
       \caption{Pearson correlation of evaluators with human scores in machine translation. \bt{bold} represents best overall scores, while \ul{underline} represents best scores within each model group (Baseline, LLaMA, Qwen). Overall, ContrastScore outperforms
       % both 
       single and ensemble methods, 
       % as well as most baseline metrics, 
       % except the state-of-the-art XCOMET-XXL with more parameters and fine-tuned on massive machine trasnlation data, 
       as well as baseline metrics for both LLaMA and Qwen families.
       % across both the LLaMA and Qwen families.
   }
   \vspace{-3mm}
   \label{tab:mt}
\end{table*}

\subsection{Bias Analysis}
%Besides the general score improvement presented in the Table ~\ref{tab:mt}, we also consider potential biases that exist in current LLMs performance. As mentioned in \S\ref{sec:meta-evalation}, we calculate the Likelihood bias and the length bias of LLMs.

Table \ref{tab:llama likelihood bias} presents the likelihood bias scores across different evaluators of the LLaMA family. 
%\xw{We compare likelihood bias within the same model family to isolate the effect of contrastive evaluation from architectural differences, enabling a clearer analysis relative to single-model and ensemble settings.} \todo{SE: an explanation why only this setup?} 
The results clearly demonstrate that ContrastScore substantially mitigates likelihood bias compared to both single-model and ensemble approaches. Specifically, ContrastScore achieves lower likelihood bias than any of the individual models employed. For instance, using LLaMA (8B, 3B), it reduces likelihood bias by 18.0\% and 64.3\% compared to the single LLaMA 8B model on QAGS-XSUM and MQM22, respectively.

ContrastScore is also effective in mitigating likelihood bias for the Qwen model family, although the improvements are somewhat smaller than those observed with LLaMA. Notably, ContrastScore consistently outperforms all individual Qwen models in reducing likelihood bias, with the exception of the 0.5B model on MQM23.
%Similarly,  ContrastScore with Qwen (7B, 0.5B) results in a reduction of 25.6\% and 34.8\% compared to single Qwen 7B model on SummEval and MQM23. 
%Similar trends are observed with the Qwen family in Table \ref{tab:qwen likelihood bias}, where ContrastScore also effectively mitigates likelihood bias.
These findings suggest that ContrastScore is highly effective in diminishing the reliance on the inherent likelihood of individual models.  See also Tables \ref{appen:likelihood_mt} and  \ref{appen:likelihood_summ} in Appendix for more details and breakdown results.

\begin{table}[t]
\centering
\resizebox{0.98\columnwidth}{!}{
\begin{tabular}{llcccc}
\toprule
\multicolumn{2}{c}{\multirow{2}{*}{\textbf{Settings}}} & \multicolumn{2}{c}{\textbf{Machine Translation}} & \multicolumn{2}{c}{\textbf{Summarization}} \\
\cmidrule(lr){3-4} \cmidrule(lr){5-6}
\multicolumn{2}{c}{} & MQM22 & MQM23 & Q-XSUM & SummEval \\
\midrule
\multirow{3}{*}{\rotatebox{90}{Single}} & 1B & 0.342 & 0.212 & 0.382 & 0.348 \\
& 3B & 0.323 & 0.245 & 0.289 & 0.385 \\
& 8B & 0.297 & 0.352 & 0.267 & 0.381 \\
\midrule
\multirow{3}{*}{\makecell[c]{\rotatebox{90}{Ensemble}}} & (3B,1B) & 0.215 & 0.123 & 0.249 & 0.308 \\
& (8B,1B) & 0.180 & 0.152 & 0.225 & 0.326 \\
& (8B,3B) & 0.222 & 0.229 & 0.242 & 0.359 \\
\addlinespace[1pt]
\midrule

\multirow{3}{*}{\makecell[c]{\rotatebox{90}{Contrast}}} & (3B,1B) & \textbf{0.058} & \textbf{0.026} & 0.233 & \textbf{0.183} \\
& (8B,1B) & 0.104 & 0.134 & 0.220 & 0.262 \\
& (8B,3B) & 0.106 & 0.137 & \textbf{0.219} & 0.240 \\
\bottomrule
\end{tabular}
}
\caption{Likelihood bias scores for machine translation and summarization tasks across LLaMA model family. 
    %In aggregated scores (ensemble and contrastive), the bias value is underscored if it is lower than the bias score of each individual model included in the aggregation.
    % The relatively highest bias value within each group is underlined. 
    The lowest overall bias score is boldfaced.
    }
    \label{tab:llama likelihood bias}
\end{table}

\begin{table}[h!]
\centering

\resizebox{0.98\columnwidth}{!}{
\begin{tabular}{llcccc}
\toprule
\multicolumn{2}{c}{\multirow{2}{*}{\textbf{Settings}}} & \multicolumn{2}{c}{\textbf{Machine Translation}} & \multicolumn{2}{c}{\textbf{Summarization}} \\
\cmidrule(lr){3-4} \cmidrule(lr){5-6}
\multicolumn{2}{c}{} & MQM22 & MQM23 & Q-XSUM & SummEval\\
\midrule
\multirow{3}{*}{\makecell[c]{\rotatebox{90}{Single}}} & 0.5B & 0.341 & \textbf{0.252} & 0.347 & 0.376 \\
& 3B & 0.442 & 0.451 & 0.349 & 0.392 \\
& 7B & 0.441 & 0.463 & 0.373 & 0.398 \\
\midrule
\multirow{3}{*}{\makecell[c]{\rotatebox{90}{Ensemble}}} & (3B,0.5B) & 0.331 & 0.327 & \textbf{0.282} & 0.379 \\
& (7B,0.5B) & 0.286 & 0.294 & 0.289 & 0.393 \\
& (7B,3B) & 0.358 & 0.381 & 0.345 & 0.369 \\
\addlinespace[1pt]
\midrule
% \vspace{1pt}
\multirow{3}{*}{\makecell[c]{\rotatebox{90}{Contrast}}} & (3B,0.5B) & 0.307 & 0.314 & 0.294 & \textbf{0.236} \\
& (7B,0.5B) & 0.287 & 0.302 & 0.318 & 0.296 \\
& (7B,3B) & \textbf{0.272} & 0.287 & 0.353 & 0.329 \\
\bottomrule
\end{tabular}
}
\caption{Likelihood bias scores for machine translation and summarization tasks across Qwen model family.
    % The relatively highest bias value within each group is underlined. 
    The lowest overall bias score is boldfaced.}
\label{tab:qwen likelihood bias}
\end{table}

\subsection{Efficiency Analysis}
\label{sec:eff}

% --
% DL: moved table and figure to appendix
% --
\begin{figure}[htbp]
  \centering
  \includegraphics[width=.49\textwidth]{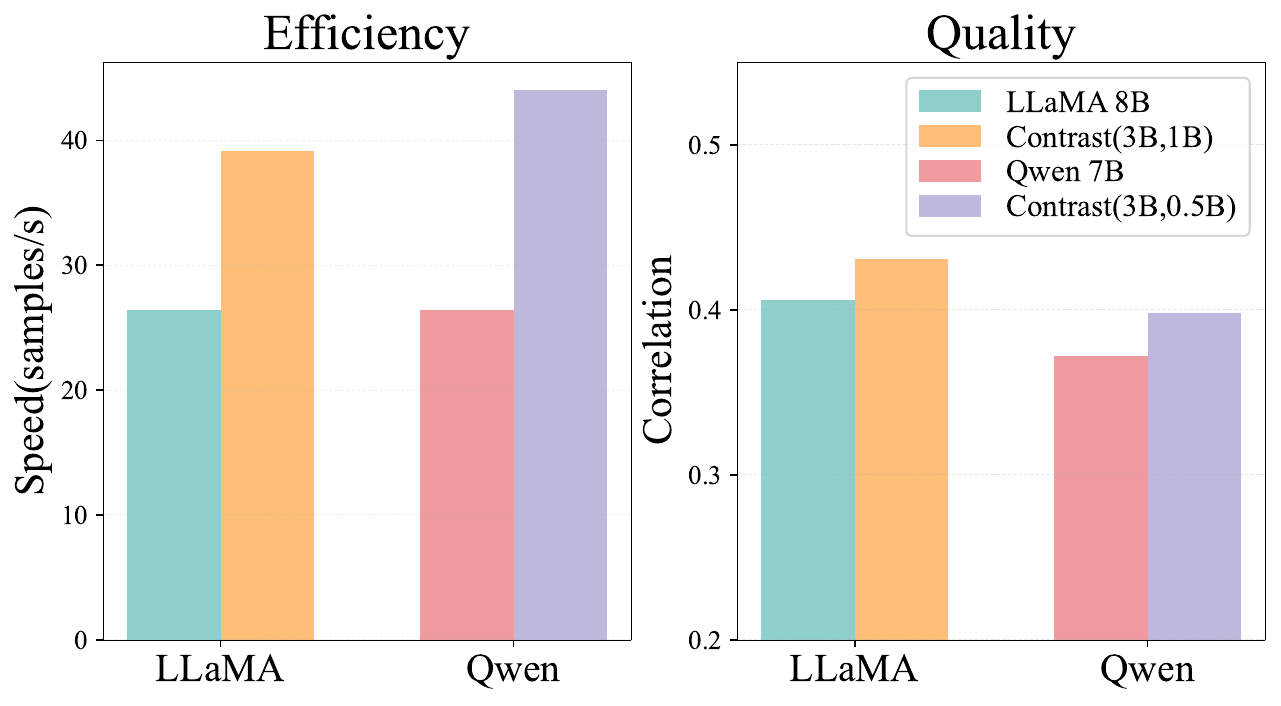} 
  \caption{Efficiency and quality of ContrastScore with smaller models compared to single larger model on summarization task. } 
  \vspace{-3mm}
  \label{fig:efficiency} 
\end{figure}

We report processing speeds for both machine translation and summarization evaluation tasks (see Table~\ref{tab:efficiency} in Appendix~\ref{appen:results}). Recall that in Tables~\ref{tab:summ} and \ref{tab:mt}, our ContrastScore employing two smaller models (for both LLaMA and Qwen families) achieves higher correlation with human judgments compared to using a single larger model in summarization, and similar trends are observed in machine translation within the Qwen family. To assess efficiency, we compare the processing speeds of ContrastScore using two smaller models with those of a single larger model. In the LLaMA family, the single large model LLaMA 8B processes 26.43 samples/s for summarization, whereas ContrastScore with LLaMA 3B and 1B achieves 39.12 samples/s, respectively---approximately 1.5$\times$ faster in both cases. Similarly, in the Qwen family, ContrastScore using Qwen 3B and 0.5B processes approximately 1.7$\times$ faster on summarization and 1.5$\times$ faster on machine translation than Qwen 7B. Figure~\ref{fig:efficiency}  further illustrates that the ContrastScore framework with two smaller models delivers improved efficiency and better evaluation quality in summarization.

\section{Further Analysis}
\label{section:A}

\iffalse
\begin{table}[t]
    \centering
    \resizebox{\columnwidth}{!}{
    \begin{tabular}{llccccc}
    \toprule[1pt]
        \multicolumn{2}{c}{\textbf{Ratio-based Score}} & \textbf{COH} & \textbf{CON} & \textbf{FL} & \textbf{RE} & \textbf{AVG}  \\ \hline
        ~ & (3B,1B) & -0.225 & -0.307 & -0.343 & -0.226 & -0.275  \\ 
        LLaMA & (8B,1B) & -0.215 & -0.262 & -0.301 & -0.181 & -0.240  \\ 
        ~ & (8B,3B) & -0.104 & -0.272 & -0.215 & -0.147 & -0.184  \\ \hline
        ~ & (3B,0.5B) & -0.220 & -0.185 & -0.249 & -0.140 & -0.198  \\ 
        Qwen & (7B,0.5B) & -0.237 & -0.147 & -0.232 & -0.125 & -0.185  \\ 
        ~ & (7B,3B) & -0.293 & -0.292 & -0.262 & -0.225 & -0.268 \\ 
        \bottomrule[1pt]
    \end{tabular}}
    \caption{The Pearson correlation of DivisionScore and human score, where the abbreviations COH, CON, FL, and RE stand for coherence, consistency, fluency, and relevance, respectively. }
    \label{tab:division}
\end{table}
\fi

\begin{table*}[htbp]
\centering
\scalebox{0.7}{
\begin{tabular}{|l|l|c|c|c|c|c|c|c|c|c|c|}
    % \hline
    \toprule
    \textbf{Source Text} & \multicolumn{11}{c|}{\begin{CJK*}{UTF8}{gbsn}产品防伪标识在哪里\end{CJK*}} \\
    % \hline
    \midrule
     & \textbf{Tokens:} & \texttt{Where} & \texttt{is} & \texttt{the} & \cellcolor{blue!25}\texttt{anti} & \texttt{-} & \texttt{fe} & \texttt{iting} & \cellcolor{blue!25}{logo}& $\Bar{\textbf{Log(P)}}$ &\textbf{Rank}\\
    % \hline
    \midrule
    \multirow{3}{*}{\makecell[l]{Hypothesis 1 \\ Human Rank: 1}} & Expert & 0.2471 & 0.7305 & 0.9922 & 0.02881 & 0.9805 & 0.7969 & 1.000 & 0.000457 & -0.717 & 2\\
    & Amateur & 2.265e-06 & 0.5625 & 0.9922 & 0.000335 & 1.000 & 1.000 & 1.000 & 0.06592 & -1.319 & 1\\
    & Contrast & 0.2471 & 0.6758 & 0.8945 & 0.02881 & 0.8789 & 0.6953 & 0.8984 & 0.006134 & -0.605 & 1\\
    \midrule
    ~ & \textbf{Tokens:} & \texttt{Where} & \texttt{is} &\texttt{the} & \texttt{anti} &\texttt{ -} & \texttt{fe} & \texttt{iting} & \texttt{product} \\
    % \hline
    \midrule
    \multirow{3}{*}{\makecell[l]{Hypothesis 2 \\ Human Rank:2}} & Expert & 0.2471 & 0.7305 & 0.9922 & 0.02881 & 0.9805 & 0.7969 & 1.000 & 0.002808 & -0.618 & 1 \\
    & Amateur & 2.265e-06 & 0.5625 & 0.9922 & 0.000335 & 1.000 & 1.000 & 1.000 & 5.841e-05 & -1.701 & 2\\
    & Contrast & 0.2471 & 0.6758 & 0.8945 & 0.02881 & 0.8789 & 0.6953 & 0.8984 & 0.002808 & -0.647 & 2 \\
    % \hline
    \midrule
    ~ &\textbf{Tokens:} & \texttt{What} & \texttt{Is} & \texttt{The} & \texttt{National} & \texttt{Debt} & \texttt{Limit} & & \\
    % \hline
    \midrule
    \multirow{3}{*}{\makecell[l]{Hypothesis 3 \\ Human Rank:3}} & Expert & 0.005951 & 0.000168 & 0.2910 & 4.268e-05 & 0.001602 & 0.004944 & & & -2.668 & 3 \\
    & Amateur & 0.000572 & 6.845e-08 & 0.1543 & 6.482e-07 & 3.123e-05 & 0.000140 & & & -4.294 & 3 \\
    & Contrast & 0.005951 & 0.000168 & 0.2754 & 4.268e-05 & 0.001602 & 0.004944 & & & -2.672 & 3\\
    % \hline
    \bottomrule[1pt]
\end{tabular}
}

\caption{Case Study: Comparison of ContrastScore with Qwen 3B as an expert and Qwen 0.5B as an amateur model for Chinese-to-English (\textbf{ZH-EN}) on MQM23. $\Bar{\textbf{Log(P)}}$ denotes mean log-probability across all target tokens.
%ContrastScore can refine the discrepancies between expert and amateur models, producing consistent ranking with human judgments.
}
\label{tab:example}
\end{table*}

\noindent\textbf{Weighted Ensemble.}~~
We investigate whether the averaged ensemble baseline used in our experiment is a sufficiently strong setup, or if a weighted combination of model probabilities could yield further improvements. Specifically, we evaluate a weighted ensemble approach that linearly combines the output probabilities of a larger and a smaller model, allowing us to examine whether tuning the weight parameter provides a meaningful advantage over uniform averaging.
\begin{equation}
p_{\text{Ens}}^t = \gamma \cdot p_{\text{EXP}}^{t}+ (1-\gamma) \cdot p_{\text{AMA}}^{t}
\end{equation}
where $\gamma \in [0,1]$ is the weight factor that controls the contribution of each model. When $\gamma = 0.5$, this formulation corresponds to that reported in our main results. To analyze the impact of $\gamma$, we conduct a parameter sweep on the summarization task using LLaMA (3B, 1B), varying $\gamma$ from 0 to 1 in increments of 0.05. The results, presented in Figure~\ref{fig:gamma}, show that the weighted ensemble achieves its highest performance at $\gamma = 0.55$, which is nearly identical to the performance at $\gamma = 0.5$ (i.e., the averaged ensemble baseline). This indicates that the simple averaged ensemble is already a strong and competitive configuration, and that further weighting does not lead to meaningful improvement.

\iffalse
\noindent\textbf{Ratio-based Contrastive Decoding Formulation.}
To demonstrate that ContrastScore's discrepancy-based probability formulation is better suited for evaluation tasks than ratio-based contrast decoding formulation in \citet{li-etal-2023-contrastive}, we test Eq.~\eqref{eq:cdscore} on the summarization evaluation task. Table \ref{tab:contrast-score-lietal} presents the Pearson correlation results between the ratio-based formulation in \citet{li-etal-2023-contrastive} and human scores for both the LLaMA and Qwen model families. In all cases, the ratio-based scores are inferior to ContrastScore. This can be attributed to the division formulation, where the score approaches 1 when both the expert and amateur models assign either high or low probabilities, thereby reducing the distinguishability for evaluation. 
\fi

\paragraph{Case Study.}
To provide a qualitative analysis, Table~\ref{tab:example} presents a case study comparing ContrastScore with a single expert model (Qwen 3B) and an amateur model (Qwen 0.5B) for Chinese-to-English (ZH-EN) machine translation.  
It can be observed that the expert model misranks \textit{Hypothesis 1} as the second-best translation, whereas human evaluators rank it as the best (Rank 1). This misranking is largely influenced by the expert model assigning a very low probability (0.000457) to \textit{``logo''}, a key token in the translation. In contrast, the amateur model assigns a significantly higher probability (0.06592) to \textit{``logo''}, demonstrating greater confidence in its relevance. ContrastScore leverages this probability discrepancy to adjust the probability of \textit{``logo''}, thereby increasing the overall score of \textit{Hypothesis 1} and correctly ranking it as the best translation. This case illustrates how ContrastScore mitigates expert model underestimation of critical tokens, leading to improved evaluation alignment with human judgment.

In the case of \textit{Hypothesis 3}, all methods (i.e., expert, amateur, and ContrastScore) correctly rank it as the worst translation (Rank 3). This consistency occurs because both the expert and amateur models assign low probabilities across all tokens, indicating poor translation quality. Unlike in \textit{Hypothesis 1}, where significant probability discrepancies require adjustment, ContrastScore makes minimal modifications since the expert and amateur models are already in agreement in their evaluation. This demonstrates that ContrastScore does not introduce unnecessary changes when the models agree, ensuring that it only refines scores when meaningful corrections are needed.

% ARR submissionversion: Table~\ref{tab:example} shows a case study comparing ContrastScore with Qwen 3B (expert) and Qwen 0.5B (amateur) on Chinese-to-English (ZH-EN) translation from MQM23. In \textit{Hypothesis 1}, the expert misranked the best human-rated translation due to assigning low probability to the critical token \textit{``logo''}. ContrastScore corrected this by leveraging the amateur model's higher confidence, aligning better with human judgment. In \textit{Hypothesis 3}, all models correctly ranked it lowest due to uniformly low probabilities, requiring no adjustment. This illustrates that ContrastScore improves ranking when needed and stays consistent otherwise.

\section{Conclusion}

In this paper, we introduce ContrastScore, an evaluation metric that leverages structured disagreement between two language models of differing capacities. Unlike conventional LLM-based evaluation methods that rely solely on a single model’s likelihood estimates, ContrastScore incorporates a weaker auxiliary model as a contrastive signal to adjust probability scores for better alignment with human judgments.
Extensive experiments demonstrate that ContrastScore consistently achieves stronger correlation with human evaluations than both single-model and ensemble-based baselines. Furthermore, ContrastScore is computationally-efficient and can effectively mitigate likelihood bias, resulting in a more robust evaluation. 

\section*{Limitations}
This work presents the following potential limitations: (1) Our investigation is limited to the LLaMA and Qwen families of models. While ContrastScore demonstrates effectiveness within these two model families, further evaluation across a broader range of large language model architectures is necessary to establish its generalizability. (2) This work relies on token-level probabilities and therefore cannot combine models from different families due to differences in their subword tokenization. In future work, word-level probability could be explored to enable mixing across model families. 
(3) The metrics explored in this paper are not competitive to the state-of-the-art (SOTA) metrics developed for MT and summarization, such as MetricX \citep{juraska-etal-2024-metricx} or XCOMET-XXL \citep{guerreiro-etal-2024-xcomet}, which may use much larger models and/or fine-tuning on human annotations. We do not claim in this paper to beat SOTA metrics, but that \emph{contrastive} principles, involving two models, are superior to non-contrastive principles for evaluation metric design within one evaluation paradigm (BARTScore, in our case). To design competitive metrics, future work should consider involving (considerably) larger models than we did in this work, beyond 8B parameters. 

\section*{Acknowledgments}
We thank the anonymous reviewers for their constructive feedback and helpful suggestions. 
Xiao Wang is supported by a studentship funded by the Department of Computer Science, University of Manchester.
The NLLG Lab gratefully acknowledges support from the Federal Ministry of
Education and Research (BMBF) via the research
grant “Metrics4NLG” and the German Research
Foundation (DFG) via the Heisenberg Grant EG
375/5-1.

\bibliography{custom}

\appendix

\section{Datasets, Metrics and Prompts}
\subsection{Datasets}
\label{appen:dataset}
\paragraph{\underline{Machine Translation. }}  \textbf{MQM22} and \textbf{MQM23} datasets are annotated by professional translators from the WMT22~\cite{freitag2022results} and WMT23~\cite{freitag2023results} Metrics Shared Tasks, based on the Multidimensional Quality Metrics (MQM) framework~\cite{lommel2014multidimensional}.  \textbf{MQM22} covers 3 language pairs: English-German (EN-DE), Chinese-English (ZH-EN) and English-Russian (EN-RU), and comprises 1315, 1875, and 1315 segments for language pair EN-DE, ZH-EN, EN-RU respectively, with every segments including 15 system outputs.  \textbf{MQM23} covers 3 language pairs: English-German (EN-DE), Chinese-English (ZH-EN) and Hebrew-English (HE-EN), including 460, 1177, and 820 segments and 12, 15, and 12 systems for EN-DE, ZH-EN, HE-EN respectively.

\noindent\paragraph{\underline{Summarization.}} \textbf{SummEval}~\cite{fabbri2021summeval} contains 1600 model-generated summaries of CNN/DailyMail articles, each annotated by 3 experts and 5 crowd workers. It  covers aspects of coherence, consistency, fluency, and relevance.  \textbf{QAGS-XSUM}~\cite{wang2020asking} includes 239 system outputs from a fine-tuned BART on XSUM dataset,  focusing on factuality aspect.

\subsection{Baseline Metrics}
\label{appen:metrics}
\textbf{BLEU}~\cite{papineni2002bleu} is based on the precision of n-grams between the MT output and its reference weighted by a brevity penalty. 

 \textbf{CHRF}~\cite{popovic-2015-chrf} uses character n-grams instead of word n-grams to compare the MT output with the reference. 

\textbf{ROUGE }~\cite{lin2004rouge} measures the lexical overlap between the hypothesis and reference. We consider 3 variants ROUGE-1, ROUGE-2, and ROUGE-L.

\textbf{COMET}~\cite{rei-etal-2020-comet} is a learnt metric that is fine-tuned to produce evaluation scores for a given translation by comparing its representation to source and reference embeddings. 

\textbf{COMET-KIWI}~\cite{rei2022cometkiwi} is a reference-free variant of COMET for machine translation evaluation.  It uses a multilingual encoder to take only the source and system output as input and predicts a continuous quality score.

% \textbf{XCOMET-XXL}~\cite{guerreiro-etal-2024-xcomet}  is an evaluation model that is trained to identify errors in sentences along with a final quality score and thus leading to an explainable neural metric.  XCOMET integrates both sentence-level evaluation and error span detection capabilities, exhibiting state-of-the-art performance in machine translation evaluation.

\textbf{BERTScore}~\cite{zhangbertscore} leverages contextual embeddings from BERT to compare words in candidate and reference sentences using cosine similarity.

\textbf{MoverScore}~\cite{zhao-etal-2019-moverscore} measures semantic similarity between a candidate and a reference by combining contextualized embeddings with Earth Mover’s Distance.

\textbf{BARTScore}~\cite{yuan2021bartscore} is a generative metric that uses BART to evaluate the generated text by calculating the probabilities of the tokens.

\subsection{Prompts}
\label{appen:prompt}

The prompts for the summarization and machine translation tasks 
%in source-based setting 
are presented in Table \ref{tab:prompt}. 
%For the reference-based setting, the prompts are "To rephrase it" and "That is to say".

\begin{table}[h]
    \centering
    \begin{tabular}{p{2cm}p{5cm}}
    \toprule[1pt]
        \textbf{Tasks} & \textbf{Prompts} \\ \hline
        machine translation & Translate the following sentence to \{target language\}:  \\ \hline
        \multirow{4}{*}{summarization} & Write an accurate, relevant,  and coherent summary of the following texts:$\backslash$n \{Article\}$\backslash$n Summary:$\backslash$n \\
        
        \bottomrule[1pt]
    \end{tabular}
    \caption{Prompts for tasks description}
    \label{tab:prompt}
\end{table}

\subsection{Additional Details}
\paragraph{Artifact Licenses}
We use the following artifacts:
\begin{itemize}
    \item Qwen 2.5 - Qwen License Agreement
    \item LLaMA 3.1 and 3.2 - LLaMA 3.1 and 3.2 Community License Agreement
    \item COMET22 - Apache 2.0 License
    \item COMET-Kiwi - CC BY NC SA 4.0
    \item SummEval - MIT License
    \item WMT23 - MIT License
\end{itemize}
Our use of those artifacts complies with license terms and applicable intended use policies.

\paragraph{PII information in data}
We did not specifically check whether the datasets used contain PII information. However, we have noticed that WMT23 authors made an effort to mask PII information with special tags. We do not disseminate any new dataset; therefore, PII protection is out of the scope of our work.

\paragraph{Package Versions and Hardware}
We use `unbabel-comet` version 2.2.0 to run baselines for MT evaluation.
Experiments were conducted on the university SLURM computing cluster with various available GPUs. For benchmarking purposes, we used H100 GPUs with 96GB of VRAM.
\section{Analysis}

\subsection{The impact of hyperparameter gamma}
\begin{figure}[h]
  \centering
  \includegraphics[width=.48\textwidth]{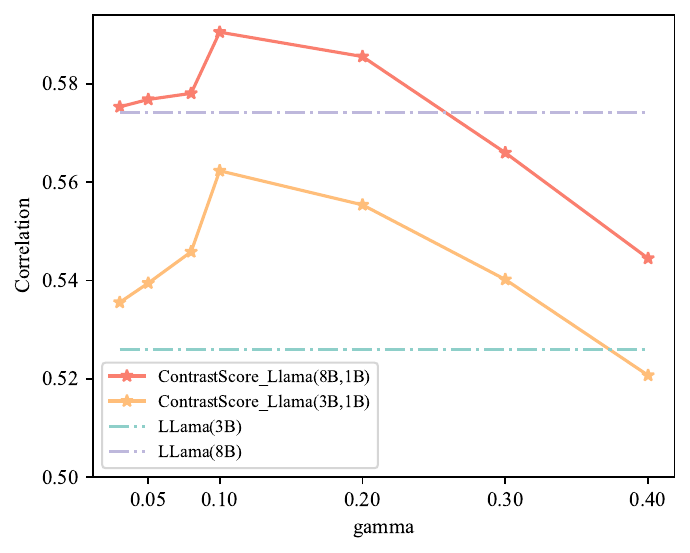} 
  \caption{Exploration of the impacts of $\gamma$. Testing correlation between evaluator score and human score in ZH-EN language pair on MQM23.} 
  \label{gamma} 
\end{figure}

We study how sensitive our method is to $\gamma$ in Figure \ref{gamma}.

We test ZH-EN language pair of machine translation task, based on LLaMA family. Figure \ref{gamma} shows that $\gamma\in[0.08,0.2]$ leads to good performance, robust in different model size settings. Furthermore,  $\gamma =0.1 $ produces the best performance, ensuring a small, controlled refinement.

\label{appen:gamma}

\subsection{Weighted ensemble}

\begin{figure}[ht]
    \centering
    \includegraphics[width=0.5\textwidth]{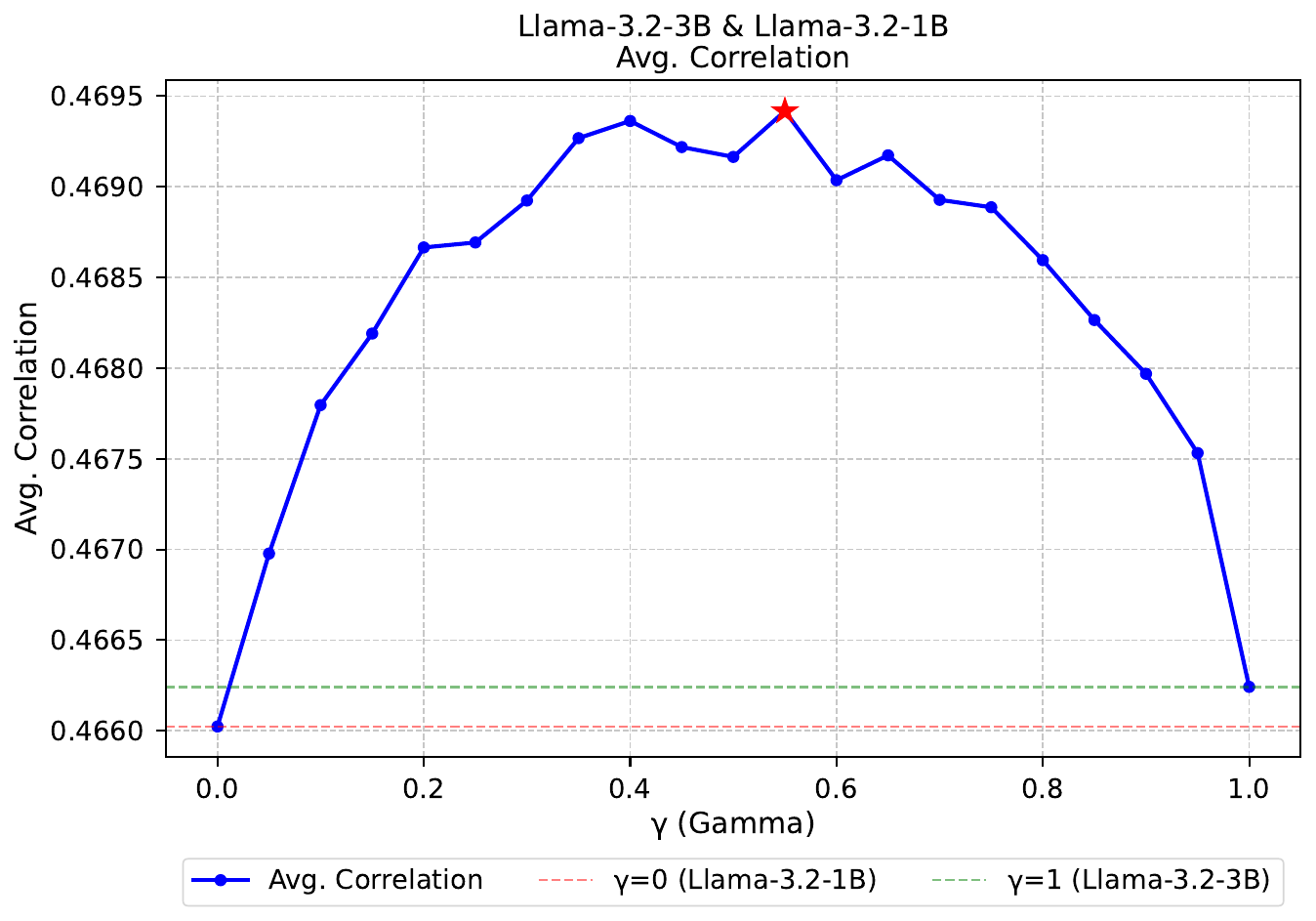}
    \caption{Exploration of weighted ensemble parameter $\gamma$ for LLaMA-3.2 (3B, 1B) on summarization. Best quality occurs at $\gamma = 0.55$ (star), but remains below ContrastScore using the same models. Horizontal lines show individual model performance. 
    %The scores reported in this ablation study differ slightly from those in Table~\ref{tab:summ} due to variations between hardware/software of experimental setups.
    }
    \label{fig:gamma}
\end{figure}

To explore whether the advantages of ContrastScore can be replicated by a simpler method based on probability combination, we evaluate a weighted ensemble approach that linearly merges the output probabilities of the larger and smaller models. The results are presented in the Figure~\ref{fig:gamma}.

% \vspace{-10pt}
\section{Results}
\label{appen:results}
\subsection{Likelihood Bias}
% Table \ref{tab:qwen likelihood bias} presents the likelihood bias scores across different evaluators of the Qwen family.
The detailed likelihood bias scores for every datasets of machine translation and summarization tasks are shown in Table \ref{appen:likelihood_mt} and Table \ref{appen:likelihood_summ}.
ContrastScore demonstrates consistent effectiveness in reducing likelihood bias across both machine translation and summarization tasks, particularly in challenging language pairs and critical evaluation aspects. In machine translation in Table \ref{appen:likelihood_mt}, ContrastScore yields substantial improvements for the LLaMA family across diverse language pairs, especially for low-resource or morphologically complex pairs such as EN-RU and HE-EN. In summarization in Table \ref{appen:likelihood_summ}, ContrastScore offers the most pronounced improvements in factuality and coherence, which are often the most difficult dimensions for language models. Overall, ContrastScore effectively mitigates likelihood bias compared to both single-model and ensemble baselines in both the LLaMA and Qwen families.

% \begin{table}[h!]
% \centering

% \resizebox{\columnwidth}{!}{
% \begin{tabular}{llcccc}
% \toprule
% \multicolumn{2}{c}{\multirow{2}{*}{\textbf{Settings}}} & \multicolumn{2}{c}{\textbf{machine translation}} & \multicolumn{2}{c}{\textbf{Summarization}} \\
% \cmidrule(lr){3-4} \cmidrule(lr){5-6}
% \multicolumn{2}{c}{} & MQM22 & MQM23 & Q-XSUM & SummEval\\
% \midrule
% \multirow{3}{*}{\rotatebox{90}{Single}} & 0.5B & 0.341 & \textbf{0.252} & 0.347 & 0.376 \\
% & 3B & 0.442 & 0.451 & 0.349 & 0.392 \\
% & 7B & 0.441 & 0.463 & 0.373 & 0.398 \\
% \midrule
% \multirow{3}{*}{\rotatebox{90}{Ensemble}} & (3B,0.5B) & 0.331 & 0.327 & \textbf{0.282} & 0.379 \\
% & (7B,0.5B) & 0.286 & 0.294 & 0.289 & 0.393 \\
% & (7B,3B) & 0.358 & 0.381 & 0.345 & 0.369 \\
% \midrule
% \multirow{3}{*}{\rotatebox{90}{Contrast}} & (3B,0.5B) & 0.307 & 0.314 & 0.294 & \textbf{0.236} \\
% & (7B,0.5B) & 0.287 & 0.302 & 0.318 & 0.296 \\
% & (7B,3B) & \textbf{0.272} & 0.287 & 0.353 & 0.329 \\
% \bottomrule
% \end{tabular}
% }
% \caption{Likelihood bias scores for machine translation and summarization tasks across Qwen model family.
%     The relatively highest bias value within each group is underlined. The lowest overall bias score is boldfaced.}
% \label{tab:qwen likelihood bias}
% \end{table}

\subsection{Efficiency}
We use processing speed as an indicator of efficiency, with detailed results for both machine translation and summarization evaluation tasks presented in Table~\ref{tab:efficiency}. Processing more samples per second indicates higher evaluation efficiency. These results demonstrate that ContrastScore offers substantial gains in evaluation efficiency, especially when leveraging smaller models.

\begin{table}[!h]
    \centering
    \resizebox{\columnwidth}{!}{
    \begin{tabular}{llcc|lcc}
    \toprule[1pt]

        ~ & \textbf{LLaMA} & \textbf{SUM} & \textbf{MT} & \textbf{Qwen} & \textbf{SUM} & \textbf{MT}  \\ \midrule
        \multirow{3}{*}{\rotatebox{90}{Single}}&1B & 63.29 & 339.04 & 0.5B & 141.96 & 552.68  \\ 
        & 3B & 47.61 & 153.25 & 3B & 48.21 & 194.22  \\ 
        &8B & 26.43 & 72.17 & 7B & 26.43 & 99.84  \\ \midrule
        \multirow{3}{*}{\rotatebox{90}{Ensemble}} & (3B,1B) & 39.51 & 110.63 & (3B,0.5B) & 44.94 & 154.5  \\ 
        & (8B,1B) & 23.81 & 66.8 & (7B,0.5B) & 26.11 & 93.92   \\ 
         & (8B,3B) & 19.81 & 53.95 & (7B,3B) & 20.35 & 71.04  \\ \midrule
        \multirow{3}{*}{\rotatebox{90}{Contrast}} & (3B,1B) & 39.12 & 109.31 & (3B,0.5B) & 44.06 & 153.99  \\ 
         & (8B,1B) & 23.33 & 66.11 & (7B,0.5B) & 25.71 & 93.41  \\ 
         & (8B,3B) & 19.05 & 53.45 & (7B,3B) & 19.61 & 70.97 \\
\bottomrule[1pt]
    \end{tabular}}   
    \caption{Processing Speed for Machine Translation and summarization evaluation tasks. Measured in Samples Per Second with batch size of 16 on single H100 GPU.}
    \label{tab:efficiency}
\end{table}

\vspace{10pt}
\begin{table*}[!h]
\centering
\scalebox{0.9}{
\begin{tabular}{llccccccccc} % Two left-aligned, eight center-aligned columns
\toprule
\multicolumn{3}{c}{\multirow{2}{*}{\textbf{Settings}}} & \multicolumn{4}{c}{\textbf{MQM22}} & \multicolumn{4}{c}{\textbf{MQM23}} \\
\cmidrule(lr){4-7} \cmidrule(lr){8-11} % Partial rules for MQM22 and MQM23 sub-headers
\multicolumn{3}{c}{} & \textbf{EN-DE} & \textbf{ZH-EN} & \textbf{EN-RU} & \textbf{AVG} & \textbf{EN-DE} & \textbf{ZH-EN} & \textbf{HE-EN} & \textbf{AVG} \\
\midrule
\multirow{9}{*}{LLaMA} & \multirow{3}{*}{Single} & 1B & 0.437 & 0.312 & 0.276 & 0.342 & 0.142 & 0.237 & 0.257 & 0.212 \\

& & 3B & 0.426 & 0.290 & 0.253 & 0.323 & 0.146 & 0.284 & 0.304 & 0.245 \\

& & 8B & 0.367 & 0.287 & 0.238 & 0.297 & 0.181 & 0.309 & 0.565 & 0.352 \\
\cmidrule(lr){2-11} % Partial rule after Single block
& \multirow{3}{*}{Ensemble} & (3B,1B) & 0.324 & 0.219 & 0.101 & 0.215 & -0.002 & 0.169 & 0.202 & 0.123 \\

& & (8B,1B) & 0.278 & 0.189 & 0.074 & 0.180 & 0.006 & 0.148 & 0.302 & 0.152 \\

& & (8B,3B) & 0.298 & 0.227 & 0.140 & 0.222 & 0.079 & 0.223 & 0.385 & 0.229 \\
\cmidrule(lr){2-11} 
& \multirow{3}{*}{Contrast} & (3B,1B) & 0.195 & 0.143 & -0.165 & 0.058 & -0.106 & 0.096 & 0.088 & 0.026 \\

& & (8B,1B) & 0.164 & 0.131 & 0.016 & 0.104 & -0.044 & 0.093 & 0.353 & 0.134 \\

& & (8B,3B) & 0.187 & 0.141 & -0.011 & 0.106 & -0.051 & 0.110 & 0.350 & 0.137 \\
\midrule 
\midrule 
\multirow{9}{*}{Qwen} & \multirow{3}{*}{Single} & 0.5B & 0.493 & 0.233 & 0.296 & 0.341 & 0.184 & 0.214 & 0.358 & 0.252 \\
& & 3B & 0.580 & 0.367 & 0.379 & 0.442 & 0.279 & 0.476 & 0.598 & 0.451 \\
& & 7B & 0.559 & 0.383 & 0.383 & 0.441 & 0.333 & 0.438 & 0.616 & 0.463 \\
\cmidrule(lr){2-11} % Partial rule after Single block
& \multirow{3}{*}{Ensemble} & (3B,0.5B) & 0.463 & 0.275 & 0.256 & 0.331 & 0.115 & 0.370 & 0.498 & 0.327 \\
& & (7B,0.5B) & 0.421 & 0.236 & 0.202 & 0.286 & 0.095 & 0.304 & 0.483 & 0.294 \\
& & (7B,3B) & 0.490 & 0.303 & 0.280 & 0.358 & 0.206 & 0.376 & 0.562 & 0.381 \\
\cmidrule(lr){2-11} % Partial rule after Ensemble block
& \multirow{3}{*}{Contrast} & (3B,0.5B) & 0.432 & 0.257 & 0.233 & 0.307 & 0.087 & 0.350 & 0.505 & 0.314 \\
& & (7B,0.5B) & 0.415 & 0.239 & 0.206 & 0.287 & 0.092 & 0.296 & 0.518 & 0.302 \\
& & (7B,3B) & 0.395 & 0.235 & 0.187 & 0.272 & 0.091 & 0.292 & 0.499 & 0.287 \\
\bottomrule
\end{tabular}
}
\caption{Likelihood bias of machine translation task for the LLaMA and Qwen family. ContrastScore can effectively mitigate the likelihood bias compared to both single and ensemble methods on MQM22 and MQM23 datasets. }
\label{appen:likelihood_mt}
\end{table*}

\begin{table*}[!h]
\centering
\scalebox{0.9}{
\begin{tabular}{llcccccccc} 
\toprule
\multicolumn{3}{c}{\multirow{2}{*}{\textbf{Settings}}} & \textbf{Q-XUM} & \multicolumn{5}{c}{\textbf{SummEval}} & \\
\cmidrule(lr){4-4} \cmidrule(lr){5-10} % Partial rules for Q-XUM and SummEval headers
\multicolumn{2}{c}{} & & \textbf{Factuality} & \textbf{Coherence} & \textbf{Consistency} & \textbf{Fluency} & \textbf{Relevance} & \textbf{AVG} \\
\midrule
\multirow{9}{*}{LLaMA} & \multirow{3}{*}{Single} & 1B & 0.382 & 0.104 & 0.471 & 0.539 & 0.279 & 0.348 \\
% \cmidrule(lr){3-10}
& & 3B & 0.289 & 0.169 & 0.477 & 0.573 & 0.321 & 0.385 \\
% \cmidrule(lr){3-10}
& & 8B & 0.267 & 0.161 & 0.477 & 0.575 & 0.311 & 0.381 \\
\cmidrule(lr){2-10}
& \multirow{3}{*}{Ensemble} & (3B,1B) & 0.249 & 0.095 & 0.406 & 0.488 & 0.240 & 0.308 \\
% \cmidrule(lr){3-10}
& & (8B,1B) & 0.225 & 0.119 & 0.417 & 0.507 & 0.261 & 0.326 \\
% \cmidrule(lr){3-10}
& & (8B,3B) & 0.242 & 0.152 & 0.450 & 0.538 & 0.295 & 0.359 \\
\cmidrule(lr){2-10}
& \multirow{3}{*}{Contrast} & (3B,1B) & 0.233 & -0.008 & 0.279 & 0.339 & 0.122 & 0.183 \\
% \cmidrule(lr){3-10}
& & (8B,1B) & 0.220 & 0.074 & 0.337 & 0.424 & 0.214 & 0.262 \\
% \cmidrule(lr){3-10}
& & (8B,3B) & 0.219 & 0.048 & 0.326 & 0.400 & 0.185 & 0.240 \\
\midrule % Single line to replicate the visual break
\midrule % Second Single line to create the "double line" effect
\multirow{9}{*}{Qwen} & \multirow{3}{*}{Single} & 0.5B & 0.347 & 0.139 & 0.489 & 0.567 & 0.308 & 0.376 \\
% \cmidrule(lr){3-10}
& & 3B & 0.349 & 0.173 & 0.492 & 0.570 & 0.332 & 0.392 \\
% \cmidrule(lr){3-10}
& & 7B & 0.373 & 0.185 & 0.489 & 0.584 & 0.334 & 0.398 \\
\cmidrule(lr){2-10}
& \multirow{3}{*}{Ensemble} & (3B,0.5B) & 0.282 & 0.170 & 0.461 & 0.567 & 0.318 & 0.379 \\
% \cmidrule(lr){3-10}
& & (7B,0.5B) & 0.289 & 0.194 & 0.464 & 0.582 & 0.332 & 0.393 \\
% \cmidrule(lr){3-10}
& & (7B,3B) & 0.345 & 0.151 & 0.472 & 0.560 & 0.295 & 0.369 \\
\cmidrule(lr){2-10}
& \multirow{3}{*}{Contrast} & (3B,0.5B) & 0.294 & 0.017 & 0.364 & 0.410 & 0.152 & 0.236 \\
% \cmidrule(lr){3-10}
& & (7B,0.5B) & 0.318 & 0.077 & 0.411 & 0.488 & 0.208 & 0.296 \\
% \cmidrule(lr){3-10}
& & (7B,3B) & 0.353 & 0.112 & 0.430 & 0.526 & 0.249 & 0.329 \\
\bottomrule
\end{tabular}
}
\caption{Likelihood bias of summarization task for the LLaMA and Qwen family. ContrastScore can effectively mitigate the likelihood bias compared to both single and ensemble methods on QAGS-XSUM and SummEval datasets. }
\label{appen:likelihood_summ}
\end{table*}

\begin{table*}[!ht]
    \centering
    \footnotesize
    \scalebox{0.99}{
    \begin{tabular}{lllccccccc}
        \toprule
        \multirow{2}{*}{}&\multirow{2}{*}{}&\multirow{2}{*}{Evaluators} & \multicolumn{3}{c}{MQM22} & \multicolumn{3}{c}{MQM23} & \multirow{2}{*}{AVG} \\
        \cmidrule(lr){4-6} \cmidrule(lr){7-9}
        & && EN-DE & ZH-EN & EN-RU & EN-DE & ZH-EN & HE-EN  & \\
        \midrule
    
        % \hline \hline
        \multirow{10}{*}{LLaMA} & \multirow{3}{*}{Single} & 
            1B          & 0.407 & 0.493 & 0.495 & 0.655 & 0.625 & 0.440 &      \cellcolor{blue!10}0.519 \\
        & & 3B          & 0.415 & 0.497 & 0.498 & 0.661 & 0.629 & 0.441 &   \cellcolor{blue!10}0.524  \\
        & & 8B          & 0.439 & 0.503 & 0.515 & 0.676 & 0.648 & 0.511 & \cellcolor{blue!10}0.549 \\
        \cmidrule(l){2-10}
        & \multirow{3}{*}{Ensemble} & 
            (3B,1B)     & 0.416 & 0.498 & 0.503 & 0.665 & 0.633 & 0.442 &    \cellcolor{blue!10}0.526 \\
        & & (8B,1B)     & 0.426 & 0.504 & 0.518 & 0.674 & 0.646 & 0.489 & \cellcolor{blue!10}0.543 \\
        & & (8B,3B)     & 0.430 & 0.504 & 0.515 & 0.675 & 0.647 & 0.490 &   \cellcolor{blue!10}0.544 \\ 
        \cmidrule(l){2-10}
        & \multirow{3}{*}{Contrast} & 
            (3B,1B)     & 0.435 & 0.502 & 0.509 & 0.674 & 0.639 & 0.438 &  \cellcolor{blue!30}0.533 \\
        & & (8B,1B)     & 0.449 & 0.507 & 0.528 & 0.682 & 0.649 & 0.496 &  \cellcolor{blue!30}0.552 \\ 
        & & (8B,3B)     & 0.445 & 0.509 & 0.530 & 0.683 & 0.652 & 0.496 &  \cellcolor{blue!30} \ul{0.553} \\ 
        \hline \hline
        \vspace{2pt}
        \multirow{10}{*}{Qwen} & \multirow{3}{*}{Single} 
        & 0.5B          & 0.395 & 0.500 & 0.505 & 0.650 & 0.637 & 0.451 &   \cellcolor{blue!10}0.523 \\
        & & 3B          & 0.417 & 0.513 & 0.502 & 0.665 & 0.659 & 0.477 &   \cellcolor{blue!10}0.523 \\
        & & 7B          & 0.427 & 0.516 & 0.517 & 0.665 & 0.655 & 0.494 &  \cellcolor{blue!10}0.539  \\
        \cmidrule(l){2-10} 
        & \multirow{3}{*}{Ensemble} 
          & (3B,0.5B)   & 0.414 & 0.510 & 0.511 & 0.663 & 0.656 & 0.479 &  \cellcolor{blue!10}0.546  \\
        & & (7B,0.5B)   & 0.418 & 0.512 & 0.518 & 0.665 & 0.655 & 0.491 &  \cellcolor{blue!10}0.539 \\
        & & (7B,3B)     & 0.425 & 0.518 & 0.518 & 0.671 & 0.663 & 0.497 &    \cellcolor{blue!10}0.543 \\
        \cmidrule(l){2-10}
        & \multirow{3}{*}{Contrast} 
          & (3B,0.5B)   & 0.428 & 0.517 & 0.519 & 0.675 & 0.666 & 0.482 &   \cellcolor{blue!30}0.549 \\
        & & (7B,0.5B)   & 0.435 & 0.519 & 0.530 & 0.676 & 0.666 & 0.499 &  \cellcolor{blue!30}0.548 \\
        & & (7B,3B)     & 0.435 & 0.521 & 0.531 & 0.676 & 0.666 & 0.499 &  \cellcolor{blue!30}\bt{0.554} \\ 
        \bottomrule
    \end{tabular}
    }
       \caption{Pairwise Accuracy of evaluators with human scores in machine translation. \bt{bold} represents best overall scores, while \ul{underline} represents best scores within each model group (Baseline, LLaMA, Qwen). Overall, ContrastScore outperforms
       % both 
       single and ensemble methods, 
       % as well as most baseline metrics, 
       % except the state-of-the-art XCOMET-XXL with more parameters and fine-tuned on massive machine trasnlation data, 
       as well as baseline metrics for both LLaMA and Qwen families.}
       % across both the LLaMA and Qwen families.

   \vspace{-3mm}
   \label{tab:pairAcc}
\end{table*}

% \iftaclpubformat

% \iffalse
% \onecolumn

% \appendix

\end{document}